\pdfoutput=1
\documentclass[11pt]{article}

\usepackage[preprint]{./latex/acl}

\usepackage{times}
\usepackage{latexsym}

\usepackage[T1]{fontenc}
\usepackage[utf8]{inputenc}

\usepackage{microtype}

\usepackage{inconsolata}

\usepackage{graphicx}

\usepackage{booktabs}
\usepackage{hyperref}
\usepackage{adjustbox}  % for multiline row
\usepackage{multirow}
\usepackage{multicol}
\usepackage{graphicx} % demo option creates a placeholder box
\usepackage{caption}
\usepackage{subcaption}
\usepackage{amsmath}

\title{From Word Sequences to Behavioral Sequences: Adapting Modeling and Evaluation Paradigms for Longitudinal NLP}

\author{
\vspace{2pt} Adithya V Ganesan$^{\dagger\ddagger}$ $\quad$  Vasudha Varadarajan$^{\odot}$ $\quad$ Oscar NE Kjell$^{\ddagger}$ $\quad$ Whitney R Ringwald$^{\delta}$ \\ \vspace{2pt}  \textbf{Scott Feltman}$^{\dagger}$ $\quad$ \textbf{Benjamin J Luft$^{\dagger}$} $\quad$ \textbf{Roman Kotov$^{\dagger}$} $\quad$ \textbf{Ryan L Boyd}$^{\Delta\Theta}$ $\quad$
\textbf{H Andrew Schwartz$^{\dagger\ddagger}$}  \\ \vspace{2pt}
$^{\dagger}$Stony Brook University $\quad$ $^{\ddagger}$Vanderbilt University $\;$ $^{\odot}$Carnegie Mellon University\\ \vspace{2pt} $^{\delta}$University of Minnesota Twin Cities $\quad$ $^{\Delta}$The University of Texas at Dallas $\;$ \\ $^{\Theta}$Texas Artificial Intelligence Research Institute  \\ \vspace{2pt}
\texttt{avirinchipur@cs.stonybrook.edu} $\qquad$ \texttt{hansen.schwartz@vanderbilt.edu}
}
\begin{document}
\maketitle
\begin{abstract}
While NLP typically treats documents as independent and unordered samples, in longitudinal studies, this assumption rarely holds: documents are nested within authors and ordered in time, forming person-indexed, time-ordered \emph{behavioral sequences}.
Here, we demonstrate the need for and propose a longitudinal modeling and evaluation paradigm that consequently updates four parts of the NLP pipeline: 
(1) evaluation splits aligned to generalization over people (\textit{cross-sectional}) and/or time (\textit{prospective}); 
(2) accuracy metrics separating between-person differences from within-person dynamics; 
(3) sequence inputs to incorporate history by default; and 
(4) model internals that support different \emph{coarseness} of latent state over histories (pooled summaries, explicit dynamics, or interaction-based models).
We demonstrate the issues ensued by traditional pipeline and our proposed improvements on a dataset of 17k daily diary transcripts paired with PTSD symptom severity from 238 participants, finding that traditional document-level evaluation can yield substantially different and sometimes reversed conclusions compared to our ecologically valid modeling and evaluation. 
We tie our results to a broader discussion motivating a shift from word-sequence evaluation toward \emph{behavior-sequence} paradigms for NLP.
\end{abstract}

\section{Introduction}
\label{sec:intro}

% Para 1: NLP has modeled meaning as word composition, but language is about people
NLP typically frames prediction as mapping isolated instances of language (e.g. documents, utterrances etc.) to an outcome.
However, documents and corresponding labels don't exist in a vacuum; they are generated by \emph{people} over \emph{time}.
% This person-indexed, time-ordered structure of documents is an individual's behavioral sequence and is inherent to how data is generated. 
% Our current machine learning techniques treats documents as independent instances failing to leverage its rich and inherent person-indexed, time-ordered structure.
% However, recent work increasingly targets inherently longitudinal problems, where multiple instances of writing are accrued per person~\cite{kumar-carley-2019-tree, sawhney-etal-2021-suicide, v-ganesan-etal-2021-empirical, tsakalidis-etal-2022-identifying}. 
% In these regimes, documents are generated by \emph{people} over \emph{time}, yielding person-indexed, time-ordered behavioral sequences rather than independent instances that current standard statistical techniques are designed for.
This mismatch propagates through the standard NLP pipeline: random document splits can leak person-specific signal and scramble temporal order, per-document training ignores informative history, and flattened accuracy metrics lack clear real-world generalization goals. % (learning stable between-person differences vs.\ tracking within-person change).
As a result, models can appear stronger than they really are for realistic applications and conclusions about the relationship between language and outcomes can even be flipped (e.g. see Figure~\ref{fig:metrics}).

% Para 2: Pervasiveness — mainstream NLP has this structure too
These violations of the independence assumption are not rare; they arise whenever many documents are produced or labeled by a limited set of authors~\citep{geva-etal-2019-modeling}, and are amplified 
when data is ordered in time. 
Consider two prominent examples from mainstream NLP involving psychological measurements of human preferences and values: Anthropic's HH-RLHF~\citep{bai2022training} reports 80\% of annotations from roughly 20 crowdworkers, and Value Kaleidoscope~\citep{sorensen2024value} collects 31k judgments from just 613 annotators.
In both cases, supervision comes from a small set of people over time, yet each document-annotation pair is treated as an independent draw.
Recognizing this rich person-indexed, time-ordered structure means evaluation must be explicit about what it is actually testing.

% Para 3: Evaluation axes
When documents are nested within individuals and ordered in time, evaluation must specify what is held out (people, time, or both), since each choice tests a different generalization problem.
We therefore separate two axes: 
(1)~\emph{cross-sectional generalization}: does a model generalize to 
unseen people? and 
(2)~\emph{prospective generalization}: does a model generalize forward in 
time for the same people?
These targets are not interchangeable, and collapsing them into a single random document split can yield misleading or even reversed inferences about model usefulness.
Research areas that center human as the unit of analysis, such as mental health NLP, personalization etc., largely treat such target-aligned evaluation as standard practice~\citep{salemi-etal-2024-lamp, tsakalidis-etal-2022-identifying}.

% Para 4: Verbal behavior frame + metric decomposition
These observations motivate wider adoption of the following best practices in NLP: treat documents as ordered samples of \emph{verbal behavior} emitted by individuals, not as word sequences~\citep{boyd-schwartz-2021-verbal, v-ganesan-etal-2024-text, soni-etal-2024-large}.
% Remove the line below if formal notation is better placed in Section 3:
Formally, each observation is indexed by person $i$ and time $t$, $(x_{i,t}, y_{i,t})$, with dependence induced by person-level structures (style, baseline levels) and temporal structures (autocorrelation, trends).
Even under target-aligned splits, however, a single pooled metric obscures what a model actually learned: in ``human-level'' applications~\cite{v-ganesan-etal-2021-empirical}, variation decomposes into \emph{between-person} differences and \emph{within-person} change~\citep{curran2011disaggregation, hoffman2009persons}, and flattened metrics conflate both --- rewarding models that memorize stable individual baselines while hiding whether they capture day-to-day dynamics~\citep{hamaker2015critique}.

To our best knowledge, this is the first work to systematically demonstrate in NLP that document-centric random splits and flattened metrics can yield qualitatively different and sometimes reversed conclusions relative to target-aligned evaluation.
Rather than optimizing to a single leaderboard score, we introduce a methodology that asks what model performance legitimately claims: ecologically valid splits, decomposed between/within-person metrics, and sequence modeling as a principled default.
We demonstrate these points on a rich longitudinal dataset, showing how standard NLP/ML evaluation choices can mischaracterize performance when documents are treated as independent examples.

% Para 6: Contributions
We make five contributions, organized from problem to implications:
(1) Demonstrating the problem: We show that traditional document-level evaluation can yield substantially different conclusions than evaluations tied to ecologically valid, real-world use-cases.
(2) Developing an evaluation framework: We operationalize evaluation splits that target generalization across \emph{people} (cross-sectional) and across \emph{time} (prospective), and show that these settings support qualitatively different inferences.
(3) Demonstrate need for metric decomposition: We propose reporting both \textit{between-person} and \textit{within-person} variants of standard metrics to disentangle person-level and temporal performance.
(4) Show the need to adapt modeling: Incorporating temporal context improves performance under these changes and brings out longitudinal modeling challenges.
% (5) Empirical grounding. We ground the methodology in a temporally dense, psychologically validated dataset with daily diaries and daily measures of PTSD symptom severity.
(5) Broader implications: We connect these findings to a shift from word-sequence evaluation toward \emph{behavior-sequence} paradigms for longitudinal NLP.

\section{Dataset}
\label{sec:data}

% \noindent\textbf{PTSD-STOP.}
We use the PTSD-STOP dataset~\cite{ringwald2025day} comprising dense longitudinal monitoring of PTSD symptoms with multimodal daily measurements.
Participants were recruited through Stony Brook's World Trade Center Health Program\footnote{\href{https://www.stonybrookmedicine.edu/WTC}{www.stonybrookmedicine.edu/WTC}} and completed a daily protocol for up to 90 days.
Each day, participants (i) recorded a brief video diary on a personal smart device and (ii) completed a self-report PTSD symptom questionnaire (PCL; \citealp{ruggero2021posttraumatic}).
Participants received \$5 per completed daily entry (video + survey), for up to \$450 across the study.

\begin{table}[!h]
    \centering
    \resizebox{\linewidth}{!}{
    \begin{tabular}{l|c|c|c|c}
    \toprule
    & \textbf{Total Data}
    & \begin{tabular}{@{}c@{}} \textbf{Cross-}\\ \textbf{Sectional} \end{tabular}
    & \textbf{Prospective}
    & \begin{tabular}{@{}c@{}} \textbf{Cross-Sectional}\\ \textbf{\& Prospective} \end{tabular} \\
    \addlinespace[.5ex]
    & \phantom{\includegraphics[width=3cm]{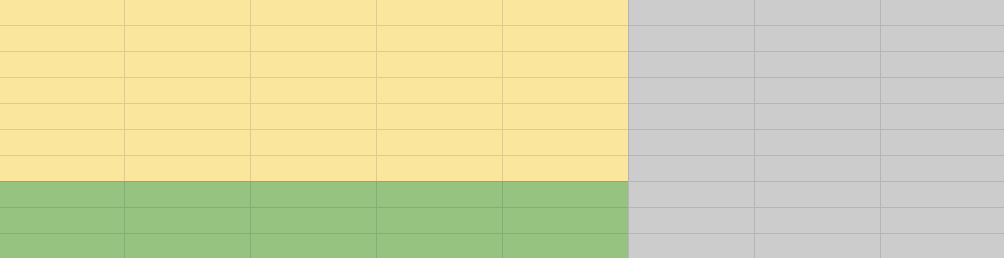}}
    & \includegraphics[width=3cm,keepaspectratio]{latex/assets/crossectionalsplit-leaf.png}
    & \includegraphics[width=3cm,keepaspectratio]{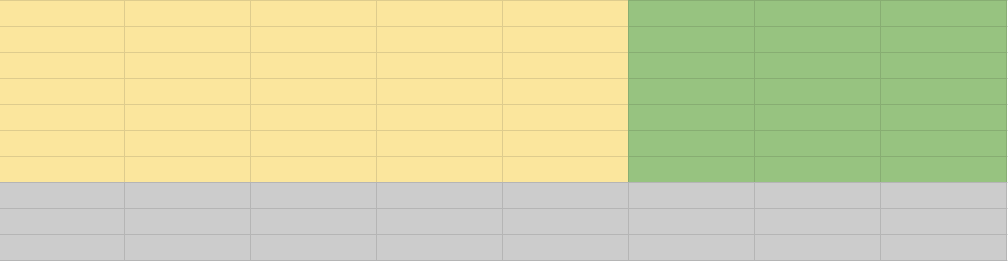}
    & \includegraphics[width=3cm,keepaspectratio]{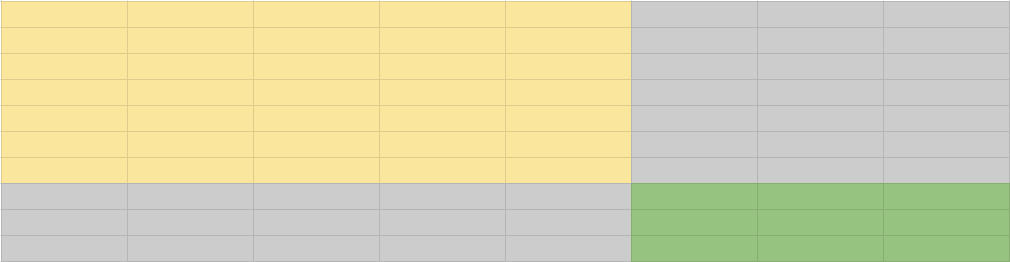} \\
    \midrule

    \# Participants & 238 & 48 & 190 & 48 \\
    \# Documents & 17{,}051 & 2{,}290 & 4{,}541 & 1{,}147 \\
    \begin{tabular}{@{}l@{}} Avg Docs per \\ Person \end{tabular}
        & 71.6 & 47.7 & 23.9 & 23.9 \\

    \bottomrule
    \end{tabular}
    }
    \caption{
    \textbf{Dataset statistics overall and within each evaluation regime.}
    Matrix icons depict how instances are partitioned for the split of interest: yellow cells indicate training person-day instances and green cells indicate the corresponding test instances (gray denotes unused).
    Statistics corresponds to the green region of respective column.
    }

    \label{tab:dataset_stats}
\end{table}

\smallskip
\noindent\textbf{Daily diary videos and transcripts.}
In the video diaries, participants described their daily experiences and emotions in response to prompts (e.g., ``\textit{tell me about the worst part of your day},'' ``\textit{describe when you felt most happy today}'')\footnote{refer \S\ref{appsec:ptsd_stop_data} for the list of prompted questions}.
We transcribe videos using the Whisper automatic speech recognition model \citep{radford-etal-whisper} and treat each transcript as a document.
The average diary length is 4.52 minutes (SD=2.12) and contains 646.6 words on average (SD=358.6).

\smallskip
\noindent\textbf{Outcome variable.}
Our primary outcome is the participant's daily PTSD symptom severity score from the PCL questionnaire, denoted $y_{it}$ for person $i$ on day $t$.
In this dataset, the PCL score ranges from 1 (low severity) to 5 (high severity).

\smallskip
\noindent\textbf{Analytic sample and coverage.}
After filtering to participants with sufficiently dense longitudinal coverage (at least 50\% days with available observations for a maximum of 90 days) and aligning diaries with outcome, our final analytic sample contains 238 participants with 17{,}051 documents (mean=71.6 documents/person).
Table~\ref{tab:dataset_stats} summarizes dataset counts overall and within the evaluation subsets used throughout the paper.

\smallskip
\noindent\textbf{Participant characteristics.}
Participants were predominantly White (60\%; 4\% Black/African American, 6\% other/multi-racial, 29\% not recorded/refused) and male (84\%), with an average age of 59.87 years (range 42--79).
We refer readers to \citet{ringwald2025day} for additional recruitment, protocol, and cohort details.

\smallskip
\noindent\textbf{Tasks.}
We study two language-to-symptom prediction settings.
For the longitudinal evaluation paradigm (\S\ref{sec:evaluation}), we consider a same-day \emph{nowcasting} task: predicting daily PTSD severity $y_{i,t}$ from same-day language $x_{i,t}$. 
For the longitudinal modeling analyses (\S\ref{sec:modeling}), we consider \emph{one-day-ahead forecasting}: predicting next-day severity $y_{i,t+1}$ from prior-day language $x_{i,t}$.

Data preprocessing steps used for creating the datasets analyzed for the evaluation paradigm (\S\ref{sec:evaluation}) is detailed in Appendix \S\ref{appsec:preprocess_long_eval} and that used for longitudinal modeling (\S\ref{sec:modeling}) is detailed in Appendix \S\ref{appsec:preprocess_long_model}.

\smallskip
\noindent\textbf{Illustration of splits.}
For ease of reading, each table/figure is accompanied by a small matrix icon indicating the evaluation regime at the top.
Rows correspond to people $i$ and columns to days $t$; each cell represents an observed instance (e.g., $(x_{i,t}, y_{i,t})$ for nowcasting, or $(x_{i,t}, y_{i,t+1})$ for forecasting).
Yellow cells denote training instances, green cells denote test instances for the evaluation split of interest, and gray cells denote instances not used for that split.
Cross-sectional evaluation splits by people ($i \in \mathcal{I}_{\text{train}}$ vs.\ $i \in \mathcal{I}_{\text{test}}$), prospective evaluation splits by time ($t \le \tau$ vs.\ $t > \tau$), and cross-sectional \& prospective splits by both (train: $i \in \mathcal{I}_{\text{train}},\, t \le \tau$; test: $i \in \mathcal{I}_{\text{test}},\, t > \tau$).

\smallskip
\noindent\textbf{Missingness handling.}
When language is missing on a given day, we impute $x_{i,t}$ by carrying forward the most recent available language observation for that participant (last observation carried forward).
In contrast, we do not impute outcomes: days with missing PCL scores ($y_{i,t}$) are excluded from training and evaluation.
Accordingly, the forecasting task uses $(x_{i,t}, y_{i,t+1})$ pairs only when $y_{i,t+1}$ is observed.

This study was approved by the Institutional Review Board of Stony Brook University and Stony Brook Medicine. 
All researchers adhered to institutional Human Subjects Research guidelines, including secure data handling (restricted access, secured storage and compute). 
All participants provided informed consent prior to enrollment, details of which are available in the original study manuscript~\cite{ringwald2025day}.

\begin{table}[!t]
    % \centering
    \resizebox{\linewidth}{!}{
    \begin{tabular}{l|c|cc}
    \toprule
    & \begin{tabular}{@{}c@{}} \textbf{Ecologically} \\ \textbf{Improbable} \end{tabular} & \multicolumn{2}{c}{\textbf{Ecologically Valid}} \\
    & \includegraphics[width=2cm,keepaspectratio]{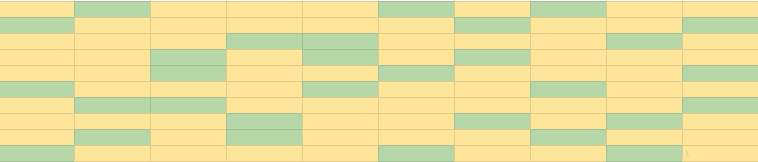} & \includegraphics[width=2cm,keepaspectratio]{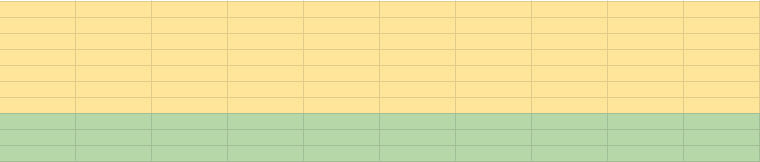} & \includegraphics[width=2cm,keepaspectratio]{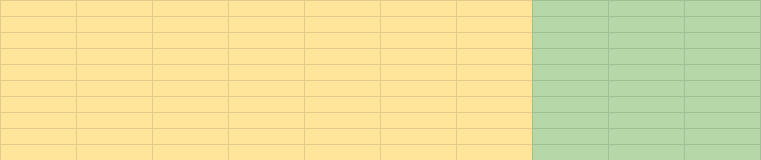} \\
    \midrule
    \begin{tabular}{@{}l@{}} \textbf{Cross Validation} \\ \textbf{Split:}  \end{tabular} & \begin{tabular}{@{}c@{}} \textbf{Traditional} \\ \textbf{Test Set} \end{tabular} & \begin{tabular}{@{}c@{}} \textbf{Cross Sectional} \\ \textbf{Test Set}  \end{tabular} & \begin{tabular}{@{}c@{}} \textbf{Prospective} \\ \textbf{Test Set} \end{tabular} \\
    & mae & mae & mae \\
    \midrule
    
    Typical model & \textbf{.520} & .757$^\ddagger$ & \textbf{.455$^\ddagger$} \\
    \begin{tabular}{@{}l@{}} Baseline: Mean of \\ Train  \end{tabular} & .660 & \textbf{.619} & .598 \\
    % \begin{tabular}{@{}l@{}} Improvement over \\  Baseline  \end{tabular} & .140 & -.139 & .144 \\
    \midrule
    N docs (train / test) & 1008 / 421 & 1008 / 421 & 1008 / 421 \\
    Mean (Train / Test)  & 1.73 / 1.78 & 1.81 / 1.60 & 1.78 / 1.67 \\
    Std Dev (Train / Test) & 0.79 / 0.84 & 0.84 / 0.71 & 0.84 / 0.72 \\
    % Variance (Train / Test) & 0.63 / 0.70 & 0.70 / 0.50 & 0.70 / 0.52 \\
    \bottomrule
    \end{tabular}
    }
    \caption{% Point 1 
    \textbf{Traditional evaluation suffers the ecological fallacy and provides no insight about generalization over people or time dimension.}
    This demonstration shows that ecologically invalid train - test split can lead to substantially different accuracies from valid settings using a typical model.
    % Point 2
    This also shows that typical models do not necessarily work well for ecologically valid settings. 
    For example, it leads to poor cross-sectional generalization as existing training methods regularize at document-level and not person-level.
    Data was sampled from 20 random people from PTSD-STOP data spanning a maximum of 90 days.
    \textit{Typical model:} RoBERTa-large with a task regression head and L2 regularization.
    $^\ddagger p<.001$ (one-sided paired t-test vs.\ baseline).
    }
    % \caption{\textbf{Random document splits can reverse conclusions under longitudinal data.}
    % On a 20-person PTSD-STOP subset, a RoBERTa-based document model outperforms a mean baseline under an ecologically improbable random split, but underperforms under a cross-sectional (unseen people) split and differs under a prospective (future time) split, despite similar label statistics across splits.
    % \textit{Typical model:} RoBERTa-large with a task regression head and L2 regularization.
    % $^\ddagger p<.001$ (one-sided paired t-test vs.\ baseline).}

    \label{tab:ecovalid}
\end{table}

\section{Evaluation: Splits and Metrics for Longitudinal NLP}
\label{sec:evaluation}

\begin{figure*}[!ht]
    \begin{subfigure}[T]{0.45\textwidth}
        \caption{\textbf{Demonstration of Flattened metric obscuring cross-sectional and temporal evaluation of the model.}}
        \label{fig:metrics_fig}
        \hfill
        \includegraphics[width=\linewidth,keepaspectratio]{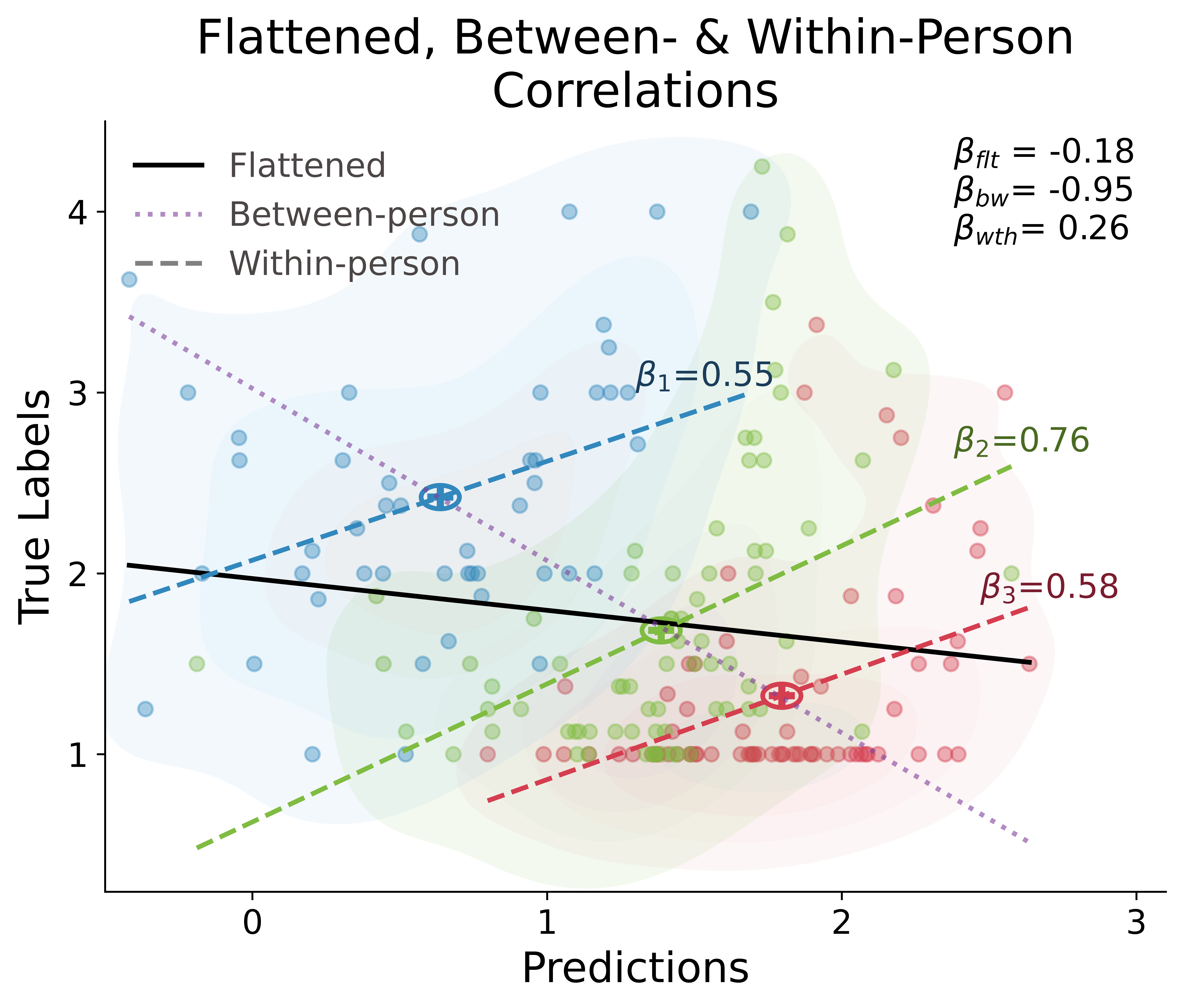}    
    \end{subfigure}
    \hfill
    \begin{subfigure}[T]{0.53\textwidth}
        \caption{\textbf{Between- and Within-Person metrics capture how well the model generalizes over people and time.}}
        \label{fig:metrics_table}
        \vspace{3ex}
        \resizebox{\linewidth}{!}{
        \begin{tabular}{l|cc|cc|cc}
        \toprule
        \begin{tabular}{@{}c@{}}\textbf{Metric}\\\textbf{Scope}\end{tabular}  & \multicolumn{2}{c|}{\begin{tabular}{@{}c@{}}\textbf{Cross-sectional}\\\textbf{Test Set}\end{tabular}} & \multicolumn{2}{c|}{\begin{tabular}{@{}c@{}}\textbf{Prospective}\\\textbf{Test Set}\end{tabular}} & \multicolumn{2}{c}{\begin{tabular}{@{}c@{}}\textbf{Cross-sectional}\\\textbf{\& Prospective}\end{tabular}} \\
        \addlinespace[.5ex]
        & \multicolumn{2}{c|}{\includegraphics[width=3cm,keepaspectratio]{latex/assets/crossectionalsplit-leaf.png}} & \multicolumn{2}{c|}{\includegraphics[width=3cm,keepaspectratio]{latex/assets/prospectivesplit-leaf.png}} & \multicolumn{2}{c}{\includegraphics[width=3cm,keepaspectratio]{latex/assets/crossprossplit-leaf.png}}  \\
        \addlinespace[.5ex]
        & MAE $(\downarrow)$ & $r (\uparrow)$ & MAE $(\downarrow)$ & $r (\uparrow)$ & MAE $(\downarrow)$ & $r (\uparrow)$ \\
        \midrule
        Flattened        & .656 & .295 & .434 & .655 & .571 & .314 \\
        \addlinespace[0.5ex]
        \hline
        \begin{tabular}{@{}l@{}}
        Between-\\Person 
        \end{tabular}    & .524 & .450 & .230 & .962 & .409 & .468 \\
        \addlinespace[0.5ex] 
        \begin{tabular}{@{}l@{}}
        Within-\\Person 
        \end{tabular}    & .654 & .285 & .426 & .297 & .578 & .214 \\
        \bottomrule
        \end{tabular}
        }
    \end{subfigure}
%     \caption{Depictions showing that traditional ML evaluation obscures the model’s ability to capture between- vs. within-person
% variation. \textit{Left:} Scatter plot of true labels (PCL scores) vs. model predictions for 3 illustrative users from the cross-sectional test set. 
%     Here, predictions covary positively within individuals over time, but the between-person fit reverses the direction of average tendencies. 
%     The traditional approach conflates these aspects, as shown by the flattened document-level fit. 
%     \textit{Right:} To address this, we propose using a separate between- and within-person metrics to assess how well the model captures individual differences and temporal variation, respectively. 
%     Using complementary metrics such as Pearson’s $r$ alongside MAE (for continuous valued variables) provides a more complete evaluation.
%     }
    \caption{\textbf{Flattened metrics can mask whether models learn people or dynamics.}
    \textit{Left:} For three illustrative users from cross-sectional test set, predictions covary within individuals, but the between-person relationship of person-level means is reversed in direction; a pooled document-level fit conflates these effects, depicted by the flattened fit.
    \textit{Right:} Decomposing performance into between-person and within-person MAE/$r$ across split regimes separates individual differences from temporal variation and clarifies what drives apparent ``good'' performance.}
    \label{fig:metrics}
\end{figure*}

% \paragraph{A model can look ``good'' or ``bad'' depending on how we split the data.}
% \paragraph{Traditional evaluation yielded substantially different errors than ecologically valid settings.}
% In longitudinal NLP datasets, documents are nested within individuals and ordered in time.
% A random document-level split can leak person identity across train and test, and can induce temporal leakage (training on later observations while evaluating on earlier ones).
% This corresponds to an interpolation-style evaluation, whereas typical behavioral and clinical deployments emphasize generalization to unseen individuals or prospective forecasting.
% Traditional evaluation setup is often easier than real use-cases that require generalization to \emph{new people} or \emph{future days}.

\paragraph{Traditional evaluation yielded substantially different errors than ecologically valid settings.}
To demonstrate the issues ensued by traditional paradigm, we mirror common settings where many documents come from a limited number of human sources \citep{geva-etal-2019-modeling} by samping a subset of 20 individuals from PTSD-STOP\footnote{details in Appendix \S\ref{appsec:preprocess_long_eval}}.
Across evaluation regimes, the \emph{same} document-level modeling pipeline led to sharply different, and sometimes reversed conclusions (Table~\ref{tab:ecovalid}).
Using a typical document-level model\footnote{Fine-tuning the task-specific/regression layer of RoBERTa-large~\cite{liu-etal-2019-roberta}. Refer to Appendix \S\ref{appsec:model_training} for details.} and a baseline that predicts the training-set mean, we find that under a \emph{traditional} random document split the model outperformed the baseline (MAE: .520 vs .660; $\Delta$MAE=$-$ .140).
However, under a more realistic \emph{cross-sectional} split (test on unseen people), this conclusion reversed: the model underperformed the baseline (MAE: .757 vs .619; $\Delta$MAE=+.138; $p<.001$).
Under a \emph{prospective} split (test on future days for the same people), the model again outperformed the baseline (MAE: .455 vs .598; $\Delta$MAE=$-$ .143; $p<.001$), but the estimated error differed substantially from the traditional split (.455 vs .520).

These discrepancies arise because longitudinal NLP data are person-indexed and time-ordered: random document splits can leak person identity across train and test and can induce temporal leakage (training on later observations while evaluating on earlier ones).
Such an interpolation-style evaluation is typically easier than deployments that require generalization to \emph{new people} or \emph{future days}.

Importantly, train/test outcome means and standard deviations were similar across paradigms (Table~\ref{tab:ecovalid}), and the baseline remained relatively stable, suggesting that the observed differences primarily reflect the evaluation protocol rather than large distribution shifts.
Overall, Table~\ref{tab:ecovalid} shows that the \emph{same} modeling pipeline can appear effective, ineffective, or even harmful depending on whether evaluation targets generalization across \emph{documents}, \emph{people}, or \emph{time}.

% To quantify this gap, we compared a typical document-level model \footnote{fine-tuning the task-specific/regression layer of RoBERTa-large~\cite{liu-etal-2019-roberta}. Refer to Appendix \S\ref{appsec:model_training} for details.} against a simple baseline that always predicted the mean label of the training set.
% Under the \emph{traditional} random document split, the typical model outperformed the baseline (MAE: .520 vs .660; $\Delta$MAE=$-$ .140).
% However, under a more realistic \emph{cross-sectional} split (test on unseen people), this conclusion reversed: the model underperformed the baseline (MAE: .757 vs .619; $\Delta$MAE=+.138; $p<.001$).
% Under a \emph{prospective} split (test on future days for the same people), the model again outperformed the baseline (MAE: .455 vs .598; $\Delta$MAE=$-$ .143; $p<.001$), but the estimated error differed substantially from the traditional split (.455 vs .520).

% Importantly, train/test outcome's means and standard deviations were similar across the three paradigms (Table~\ref{tab:ecovalid}), and the baseline remained relatively stable, suggesting that the observed differences primarily reflected the evaluation protocol rather than large distribution shifts.
% Overall, Table~\ref{tab:ecovalid} showed that the \emph{same} modeling pipeline could appear effective, ineffective, or even harmful depending on whether evaluation targeted generalization across \emph{documents}, \emph{people}, or \emph{time}.

\begin{table}[!ht]
    \centering
    \resizebox{\linewidth}{!}{
    \begin{tabular}{l|l}
    \toprule
    
    \begin{tabular}{@{}l@{}} \textbf{\textsc{Metric Scope}}\end{tabular} & \textbf{\textsc{Equation}} \\ 
    
    \midrule
    
    \textsc{Flattened} & $\frac{1}{\sum\limits_{i=1}^n t_i} \left(\sum\limits_{i=1}^n \sum\limits_{j=1}^{t_i} f(y_{ij}, \hat{y}_{ij}) \right)$ \\
    \textsc{Between-Person} & $\frac{1}{n} \sum\limits_{i=1}^n f(\bar{y}_i, \bar{\hat{y}}_i)$  \\
    \textsc{Within-Person} & $\frac{1}{n} \sum\limits_{i=1}^n \left( \frac{1}{t_i} \sum\limits_{j=1}^{t_i} f(y_{ij}, \hat{y}_{ij}) \right)$ \\
    % \textsc{Mixed} & $\sqrt{\text{between-person}^2 + \text{within-person}^2}$ \\
    \bottomrule
    \end{tabular}
}
    % \caption{
    % \textbf{Traditional Machine Learning metrics (\textsc{Flattened}) obscure evaluation of behavioral models over temporal and cross-sectional dimensions}.
    % Alternatively, Within-person metrics captures the model's temporal capabilities and Between-Person metrics captures cross-sectional (or person-level/ traitful) capabilities.
    % $f$ represents the evaluation metric, which is MAE in this case, and $y_{ij}$ and $\hat{y}_{ij}$ are the observed and predicted behaviors for person $i$ at time $t$.
    % }
    \caption{\textbf{Metric scopes for person-indexed longitudinal evaluation.}
    Flattened metrics pool all person-days, while between-person and within-person metrics separately evaluate person-level differences and within-person temporal variation.
    Here $f$ is the per-instance metric function, and $y_{ij},\hat{y}_{ij}$ are the observed and predicted outcomes for person $i$ at time $j$.}
    \label{tab:other_metrics}
\end{table}

\paragraph{A single pooled metric hid whether the model learned people or dynamics.}
Even with realistic splits, standard ``flattened'' metrics (computed by pooling all documents together) mixed two distinct prediction goals:
(i) capturing stable differences between people (who tended to have higher scores on average) and
(ii) capturing within-person change over time (day-to-day deviations).
These are both important in longitudinal behavioral prediction, but a model can perform well on one while performing poorly on the other.
We therefore propose reporting complementary \emph{between-person} and \emph{within-person} metrics (Table~\ref{tab:other_metrics}).
Between-person metrics evaluate predictions after aggregating within each person (e.g., comparing each person’s mean predicted score to their mean true score), while within-person metrics evaluate performance within each person first and then average across people.
In simple terms, between-person evaluates how well the model captured \emph{who was higher}, and within-person evaluates how well it captured \emph{when someone was higher}.

\paragraph{Between- and within-person metrics reveal what drives ``good'' performance.}
Figure~\ref{fig:metrics_fig} illustrates why pooled evaluation could be misleading.
For three example individuals from the cross-sectional test set, predictions covaried with the true labels within each person over time, yet the relationship between person-level averages differed in direction in this illustrative subset.
When documents were pooled, these sources of variation were blended into a single summary that obscured which component the model captured.

Figure~\ref{fig:metrics_table} makes this distinction explicit by decomposing performance into between-person (individual differences) and within-person (temporal dynamics) components.\footnote{We resample the data to include the third evaluation split; details are in Appendix \S\ref{appsec:preprocess_long_eval}.}
Across all regimes, between-person performance exceeds within-person performance.
On the \emph{cross-sectional} test set, between-person error and correlation are better than within-person (MAE: .524 vs .654; $r$: .450 vs .285), indicating that the model more reliably captures stable person-level differences than day-to-day variation.
The gap is larger on the \emph{prospective} test set: between-person correlation is near-perfect ($r=.962$; MAE=.230) while within-person correlation remains modest ($r=.297$; MAE=.426), implying that the strong flattened prospective correlation (flattened $r=.655$) is driven primarily by person-level signal.
The combined cross-sectional \& prospective setting shows the same pattern (between-person $r=.468$ vs within-person $r=.214$).
Overall, reporting between- and within-person metrics alongside pooled metrics provides a clearer account of model behavior in longitudinal NLP: whether performance is driven by learning \emph{who} differs or by tracking \emph{how} individuals change over time.

\section{Modeling: From Isolated Documents to Behavioral Sequences}
\label{sec:modeling}

% \paragraph{Traditional models perform best on Cross-sectional sets at lower dimension sizes and on Prospective set at higher dimension sizes.} 
% \subsection{Modeling: from isolated documents to behavioral sequences}
\begin{figure}[!ht]
    \centering
    \includegraphics[width=\linewidth]{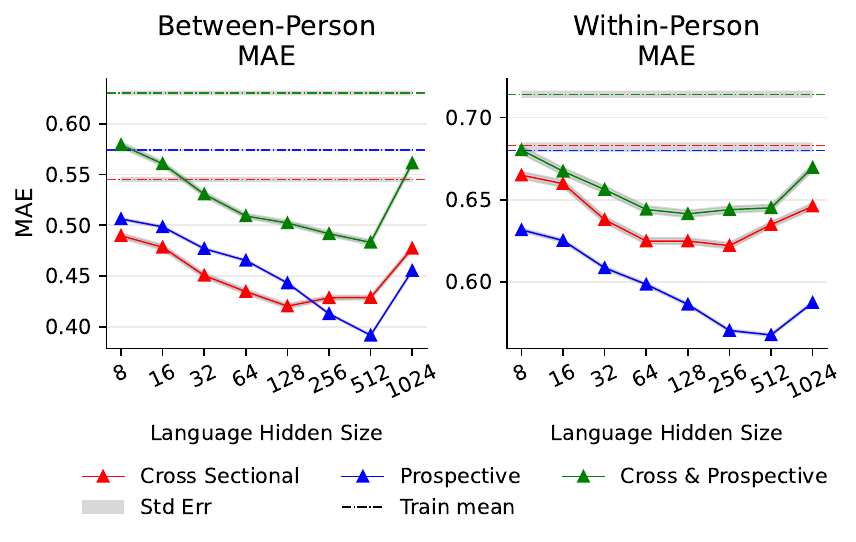}
    \caption{\textbf{Between- and within-person MAE vs.\ representation size.}
    Error shows a U-shaped trend across regimes: cross-sectional and cross-sectional \& prospective perform best with small representations (64--128), whereas prospective benefits from larger size ($\sim$512).
    % Overall, generalizing to unseen people is harder than forecasting forward in time.
    }

    \label{fig:mae_vs_hdnsize}
\end{figure}

\begin{figure*}[!h]
    \centering
    \includegraphics[width=\linewidth]{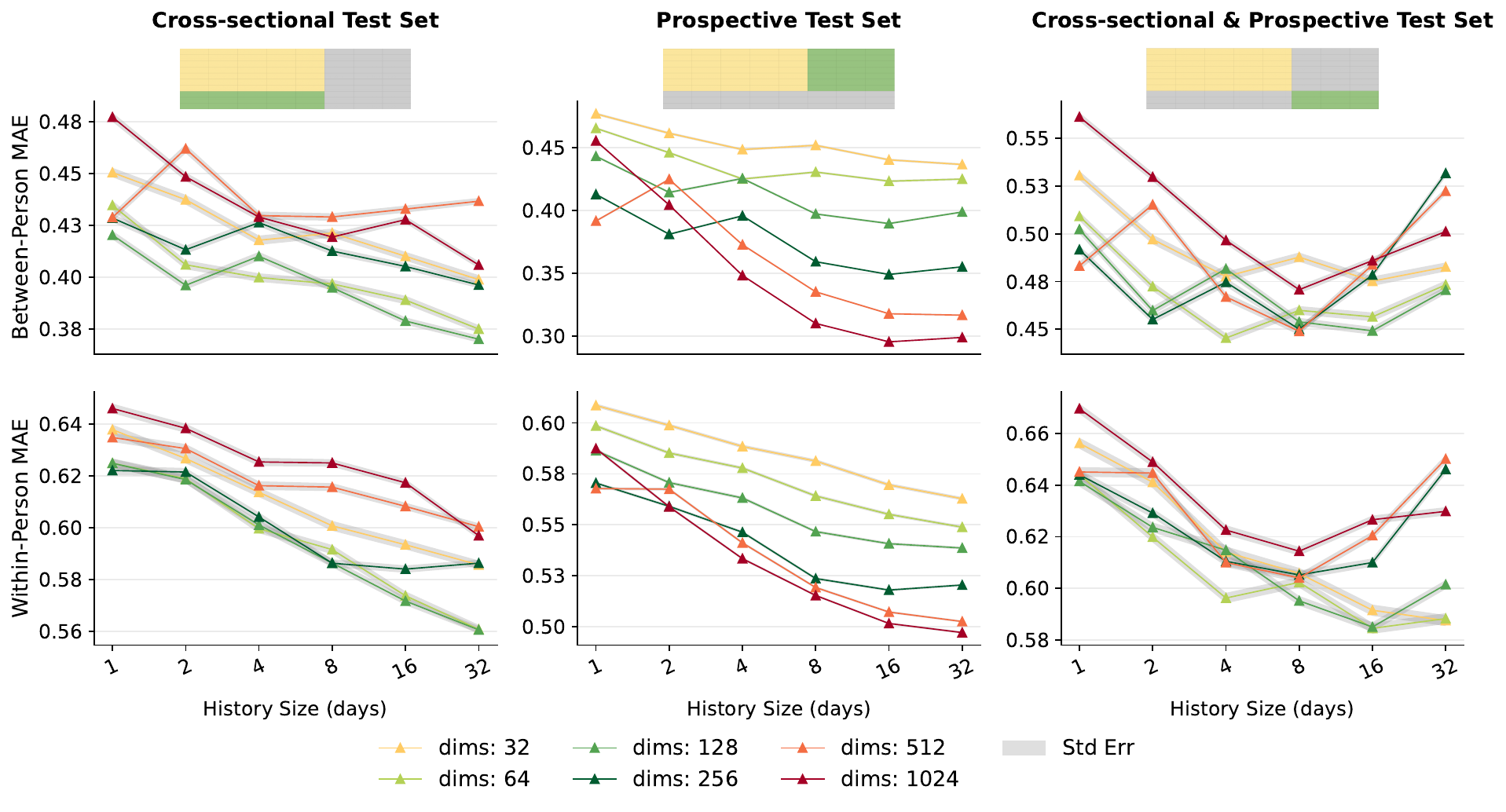}
    \caption{\textbf{Between- and within-person MAE vs.\ history length.}
    Using longer history generally improves performance, but the best history--capacity trade-off depends on the regime: prospective benefits from longer context with larger representations ($\geq$512), while cross-sectional and cross-sectional \& prospective improve primarily with longer context at smaller size ($\sim$64).
    Across splits, the average best size is 128.}
    \label{fig:mae_vs_history}
\end{figure*}

\paragraph{Representation capacity depended on the generalization target (people vs.\ time).}
We first considered the typical NLP modeling setup that maps an \emph{isolated} document representation to an outcome, and varied the dimensionality of the language representation (``hidden size'').
Prior work in human-level NLP suggested that relatively low-dimensional representations can suffice for prediction \citep{v-ganesan-etal-2021-empirical}; applying our evaluation paradigm added an important nuance.
We reduced the dimensions of language using Principal Component Analysis. 

As shown in Figure~\ref{fig:mae_vs_hdnsize}, the optimal dimensionality shifted depending on whether evaluation required generalization to \emph{new people} (cross-sectional), to \emph{future days} (prospective), or both (cross-sectional \& prospective).
In the cross-sectional and cross-sectional \& prospective test sets, performance was best at relatively low dimensionality (64-128), consistent with prior qualitative finding that increasing representation size does not monotonically improve human-level prediction.
In contrast, for the prospective test set, error decreased further at higher dimensionality (around 512), and this pattern held for both between-person and within-person MAE.

Across settings, prospective errors were lower than cross-sectional errors, indicating that generalizing to \emph{unseen people} was more difficult than generalizing forward in \emph{time} for seen people.
Finally, all language-based models outperformed the ``mean of train'' baseline (dotted lines), and the same qualitative trends were observed under SMAPE (Figure~\ref{fig:smape_vs_hdnsize}).

\paragraph{Modeling language as behavioral \emph{sequences} improves prediction under ecologically valid evaluation.}
We next tested whether treating daily diaries as a \emph{sequence of behaviors}, rather than independent documents, improves performance.
Using an autoregressive ridge model, we varied the history length $h$ (number of prior days of language representations provided as input), where $h{=}1$ corresponds to the traditional setup.

As shown in Figure~\ref{fig:mae_vs_history}, incorporating longer histories reduced error across evaluation sets, with the most consistent gains in the cross-sectional split, where MAE improved monotonically with $h$.
In the prospective split, autoregressive performance improved initially and then plateaued as history increased, while in the cross-sectional \& prospective split, autoregressive performance exhibited a U-shaped relationship with $h$, suggesting an accuracy -- complexity trade-off under the most stringent generalization setting.

Across splits, the best-performing sequence models achieved improvements that exceeded the uncertainty in our MAE estimates (standard errors were non-overlapping at the best-performing settings), indicating that the gains are robust and not driven by noise.
Together, these results motivate \emph{sequence modeling of language as the default} for longitudinal behavioral prediction.

\begin{figure*}
    \includegraphics[width=\linewidth]{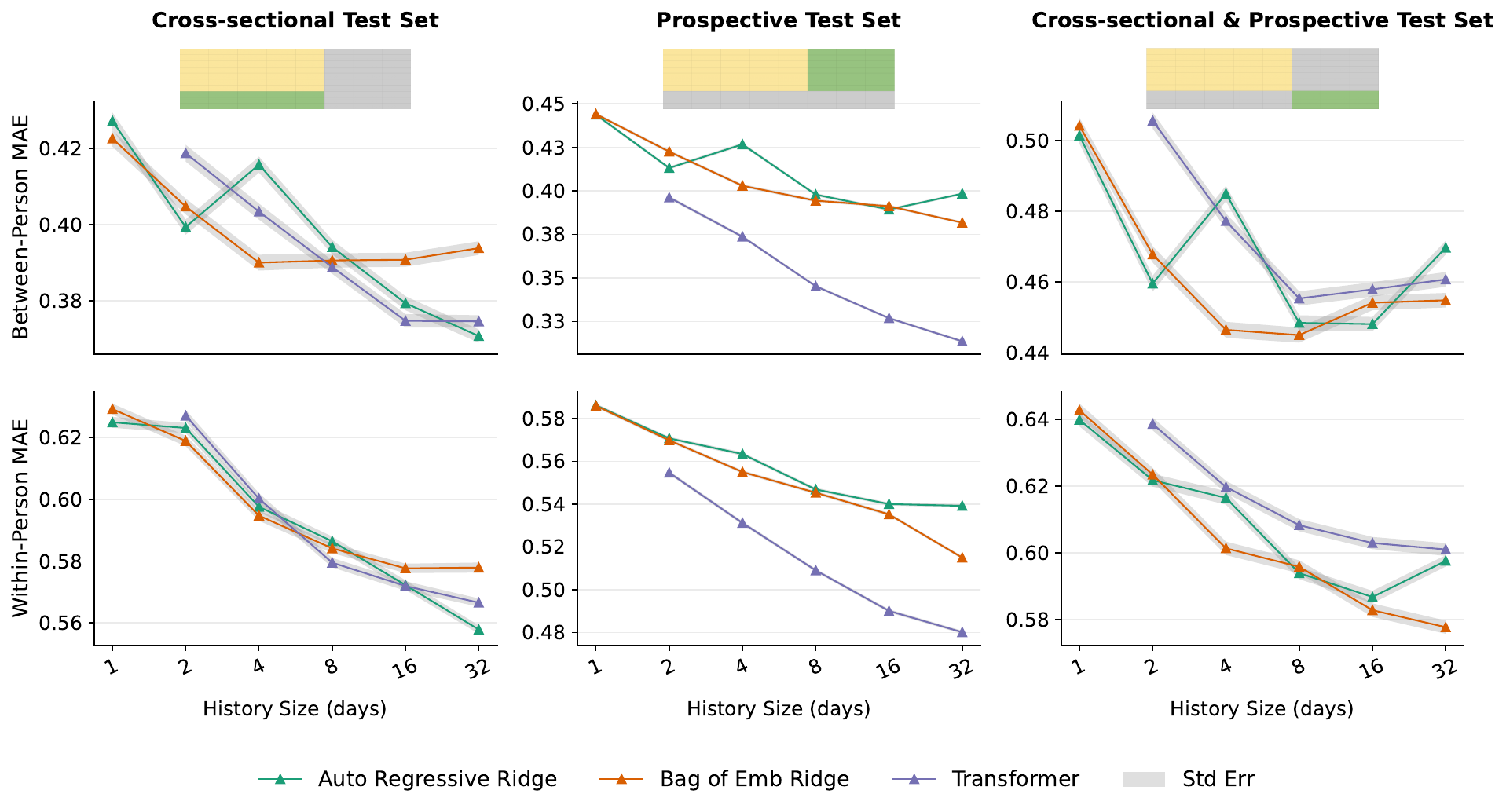}
    \caption{\textbf{AR vs.\ BoE vs.\ Transformer across history length (128 dims per day).}
Between-person (top) and within-person (bottom) MAE as a function of history $h$.
AR performs best for cross-sectional generalization, while modeling temporal interactions (Transformer) yields the largest gains for prospective generalization but performs worst for cross-sectional \& prospective.}
    \label{fig:mae_vs_history_ARBoETrns}
\end{figure*}

\paragraph{The optimal temporal inductive bias depends on the generalization target, implying a need for \emph{state coarseness}.}
To understand what aspects of history modeling drive these gains, we compared three increasingly expressive ways to map a length-$h$ language history into a predictive state: (i) a \emph{bagged history} (BoE) that averages the last $h$ days' representations into a single coarse summary vector, (ii) an \emph{autoregressive} model (AR) that explicitly parameterizes lagged dynamics over the $h$ days, and (iii) a small \emph{transformer} that can model content-dependent \emph{interactions} across days (trained from scratch with a causal mask limiting attention to the past $h$ days; Figure~\ref{fig:mae_vs_history_ARBoETrns}).

Strikingly, the best-performing inductive bias depended on the evaluation split.
In the cross-sectional split, AR performed best, with the transformer close behind, suggesting that structured dynamics help while more flexible interaction modeling provides limited additional benefit.
In the prospective split, the transformer clearly outperformed AR and BoE, consistent with the presence of predictive signal in temporal interactions once individual baselines are observed.
In the cross-sectional \& prospective split, BoE performed best, indicating that a \emph{coarser}, more regularized state is advantageous under the most stringent generalization setting.

Overall, no single temporal inductive bias dominates; instead, models benefit from an internal structure that can accommodate different \emph{levels of state coarseness}, ranging from pooled summaries, to explicit dynamics, to interaction-rich sequence models, depending on whether the task requires generalization across people, time, or both.

% \paragraph{Modeling temporal interactions of behavioral representations.}
% To test if modeling temporal interactions of behavioral representations produce more accurate predictions, we train the simplest transformer model from scratch (1 layer, 1 head, no positional embeddings) on the behavioral sequences.
% In order to keep it directly comparable to the auto-regressive ridge results above, we use a lower-block diagonal masking on self-attention matrix to limit the number of attended history to `\textit{h}' days. 

% With ${1/5}^{th}$ hidden dimensions (128), transformers on the prospective set eclipsed the best performing auto regressive model on both between- and within-person metrics (c.f. 1024 dimensional model from Figure \ref{fig:mae_vs_history} and Transformer model from Figure \ref{fig:mae_vs_history_ARBoETrns}). 
% However, on cross-sectional set, it performed slightly worse than its auto regressive model for the longest history size (32 days), and much poorer than bag of embeddings model on cross-sectional \& prospective split.
% This result highlighted that transformers show a stark improvement over simpler approaches for seen users but performs just as good (cross sectional) or worse (cross sectional \& prospective) than auto regressive modeling.  
% % Taken together, it shows the challenge in getting complex architectures like transformers to generalize to unseen people.

\section{Discussion}

% Discussion Para 1: The i.d. assumption is (implicitly) a human-source assumption
\paragraph{The independence assumption is (implicitly) a human-source assumption.}
Our results show that assuming documents are independent is rarely an inert or inconsequential abstraction in NLP; it implicitly assumes the \emph{human sources} that generate supervision (authors, annotators) are exchangeable and independent.
Because supervision comes from a finite set of people, models can exploit source-specific regularities and fail to generalize to unseen sources, underscoring the importance of documenting dataset provenance \citep{geva-etal-2019-modeling, bender-friedman-2018-data}.
This aligns with human-centered NLP that treats language as \emph{verbal behavior} produced by people in context \citep{boyd-schwartz-2021-verbal, soni-etal-2024-large}.
Our evaluation framework operationalizes this perspective by making the intended generalization target explicit: \emph{new documents from seen people}, \emph{new people}, and/or \emph{future time}.

% Discussion Para 2: Ecological validity as distribution shift + what splits are for
\paragraph{Model evaluations lose meaning and value when they lack veridicality.}
Our cross-sectional and prospective splits instantiate evaluation under distribution shift: performance depends on whether test-time data share the same sources (people) and time period as training, and practical deployments often restrict this overlap (e.g., predicting for \emph{unseen individuals} or \emph{future days}).
This framing aligns with in-the-wild robustness work and evidence that models can learn ``shortcuts'' that look strong under convenient test sets but break under shifted conditions \citep{koh-etal-2021-wilds, geirhos-etal-2020-shortcut, rosenblatt2024data}.
In longitudinal NLP, temporally aware evaluation has been motivated as necessary for understanding behavioral dynamics rather than static aggregates \citep{matero-schwartz-2020-autoregressive, tsakalidis-etal-2022-identifying}.
The same logic applies, perhaps with greater urgency, to mainstream NLP settings where person-indexed structure is present but unacknowledged: RLHF pipelines and value annotation data where small, repeated contributor pools are evaluated as if each annotation is independent but the leakage and conflation problems we demonstrate here still apply.
Our findings reinforce this methodological point for NLP: \textit{evaluation design is not merely an implementation choice; it changes the scientific claim a paper is able to support.}

% Discussion Para 3: Metrics as part of the target; pooled scores can mislead
\paragraph{Metrics should reflect whether we care about \emph{people}, \emph{time}, or both.}
When documents are repeated measures of individuals, a single pooled (flattened) score can misalign with the intended behavioral claim by mixing stable between-person differences with within-person change (cf.\ ecological correlations \citealp{robinson-1950-ecological}).
Reporting \emph{between-person} and \emph{within-person} variants of standard metrics makes these targets explicit and helps diagnose whether a model primarily recovers stable baselines versus tracks day-to-day variation.
This mirrors measurement practice in psychology and longitudinal designs, where separating within- and between-person signal is foundational to interpretation \citep{shiffman-etal-2008-ema}.
For NLP researchers, the practical recommendation is lightweight: keep standard pooled metrics for comparability, but add target-aligned decompositions when the data are person-indexed over time.

% Discussion Para 4: Modeling implications + data collection/reporting recommendations
\paragraph{Implications for modeling and dataset construction.}
Treating documents as behavioral sequences surfaces modeling choices that independence evaluation can hide, including conditioning on temporal context  to model states and trajectories rather than isolated  utterances~\citep{matero-etal-2021-melt-message, soni-etal-2022-human}.
Although human-centered tasks increasingly highlight longitudinal structure and temporally sensitive evaluation, these practices have not yet become default in mainstream NLP benchmarks~\citep{tsakalidis-etal-2022-identifying, 
matero-schwartz-2020-autoregressive}.
We therefore recommend a reporting norm for datasets with finite human sources or repeated measures: describe source structure, justify the generalization target, and align splits and metrics to that target~\citep{bender-friedman-2018-data}.
Critically, this requires that datasets preserve and disseminate  the person-indexing and time-ordering inherent to how data was collected: once these are stripped or obfuscated, downstream researchers cannot recover the structure needed for target-aligned evaluation, even if they want to.

\paragraph{Missingness as behavioral signal.}
A further implication concerns missingness. 
In longitudinal behavioral data, non-response is rarely random and it can 
itself be a signal correlated with the target state.
Work in clinical prediction has shown that explicitly modeling missingness via binary indicators improves performance~\citep{lipton2016directly}, and that imputation-aware architectures further reduce bias under irregular sampling~\citep{che2018recurrent}.
Longitudinal NLP datasets should therefore not discard missingness patterns but report and, where possible, model them.

Taken together, these steps move longitudinal NLP toward a methodology where conclusions are about \emph{people and time} --- not only about fitting collections of documents.

\section{Related Work}

\paragraph{Target-aligned evaluation for longitudinal NLP.}
A growing body of work recognizes that many applied NLP settings require temporally sensitive evaluation, especially in mental health and affective modeling \citep{matero-schwartz-2020-autoregressive, tsakalidis-etal-2022-identifying}. 
However, longitudinal evaluations often operationalize only one axis at a time (e.g., within-user temporal change or user-level holdout), making it hard to interpret what kind of generalization a reported score actually reflects. 
Target-aligned evaluation splits, using person-disjoint or time-disjoint splits, has been operationalized in the personalization literature~\citep{welch-etal-2022-leveraging, mireshghallah-etal-2022-quantifying, salemi-etal-2024-lamp}, and ecological validity has been invoked as a criterion for benchmark construction in dialogue~\citep{de2020towards} and in mental health NLP to motivate person-centered modeling~\citep{soni-etal-2022-human, jiang-etal-2023-ecologically}.
Our work complements this line by explicitly separating \emph{cross-sectional}, \emph{prospective}, and \emph{cross-sectional \& prospective} regimes and, to our knowledge uniquely, empirically demonstrating how each regime produces different conclusions from the same modeling pipeline.

\paragraph{Between-person vs.\ within-person variation.}
In ``human-level'' prediction, several studies emphasize that the effective sample size is the number of \emph{people} rather than \emph{documents}, and that performance is strongly shaped by person-level variation \citep{v-ganesan-etal-2021-empirical}. 
Conversely, longitudinal NLP work has highlighted within-person change as a core target\citep{tsakalidis-etal-2022-identifying}. 
To our knowledge, prior work rarely unifies these two sources of variation into a single evaluation framework; our split regimes and between-/within-person metrics are designed to jointly diagnose generalization across people \emph{and} across time.

\paragraph{Source-specific leakage goes beyond coarse human attributes.}
Related concerns appear in work on non-i.i.d.\ supervision, showing that models can exploit source-specific regularities and fail to generalize when the human source changes. 
For example, \citet{geva-etal-2019-modeling} demonstrate that models can leverage annotator identifiers and that performance drops under annotator-disjoint splits, motivating source-aware evaluation. 
More recently, \citet{orlikowski-etal-2023-ecological} show that even when sociodemographic attributes correlate with label variation, modeling those attributes does not recover individual annotator behavior, highlighting that leakage cannot be reduced to coarse metadata; our setting instantiates an analogous issue for \emph{authors} with the added complication of temporal ordering.

\paragraph{Modeling people over time.}
A parallel line of work adapts models to individuals via user representations or personalization, and human-language modeling arguments similarly treat individuals as generators of text with states and traits \citep{lynn-etal-2017-human, soni-etal-2022-human}. 
In longitudinal prediction tasks, autoregressive and sequence-aware models have been used to exploit temporal context \citep{matero-schwartz-2020-autoregressive}.
Related work in temporal user modeling has tracked how user-level language patterns shift over time~\citep{amir2017quantifying} and examined how models can be adapted to capture individual longitudinal trajectories~\citep{hills-etal-2023-creation}.
Our contribution is complementary: we show that making the evaluation target explicit changes which modeling choices look favorable, motivating sequence inputs by default and model internals that support different coarseness of state representations depending on the regime.

\section{Conclusion}
We argue that traditional NLP evaluation practice implicitly assumes independence between documents, even when datasets are generated by a finite set of humans and contain repeated measures over time.
Using a temporally dense daily-diary dataset, we showed that this random document-level splits can yield substantially different, and sometimes reversed conclusions relative to ecologically valid evaluation targets.
To make evaluation claims explicit, we operationalized split paradigms that separately test generalization to unseen people (cross-sectional) and to future days (prospective).

We further showed that standard pooled metrics can hide what a model actually learned in longitudinal settings, and that reporting between-person and within-person metrics clarifies whether performance is driven by stable individual differences or sensitivity to day-to-day change.
Finally, we connected evaluation to modeling: once the target is specified, incorporating temporal context becomes a principled modeling choice, and its benefits (and failure modes) become measurable.
Overall, our goal is not a benchmark race on a single endpoint, but a practical methodological shift toward longitudinal NLP as modeling \emph{people and time}, with evaluation protocols that support the scientific and deployment claims we want to make.

\section*{Limitations}

% Limitations Para 1: Methodological scope (single outcome; limited model classes; generalization of claims)
\paragraph{Scope of methods and models studied.}
This paper is a methodology paper rather than a benchmark race, and our experiments intentionally focus on simple, transparent model classes (ridge regression and autoregressive variants) to isolate evaluation effects.
As a result, we do not claim that the specific optimal representation dimensionalities or history lengths we observed will transfer unchanged to other datasets, tasks, or model families.
We leave such line of work~\cite{v-ganesan-etal-2021-empirical, singh-etal-2025-systematic} to future works.
More expressive sequential architectures, additional modalities, or alternative training objectives may change absolute performance~\cite{matero-etal-2021-melt-message, rao-etal-2025-whispa}, but the core risk we highlight—that convenient independence-assumed evaluation can reward reliance on source-specific regularities and fail under shifted regimes—has been observed broadly in ML under distribution shift and shortcut learning \citep{koh-etal-2021-wilds,geirhos-etal-2020-shortcut}.
Evaluating modern instruction-tuned LLMs under the same target-aligned splits is an immediate open direction; we note that the statistical leakage risk applies equally or more acutely to over-parameterized models capable of memorizing user-specific behavioral patterns.
Finally, we study one clinical endpoint (daily PCL) as a concrete case; extending this framework to other outcomes (e.g., functional impairment, treatment response) and to other longitudinal NLP settings is an important direction for future work.

% Limitations Para 2: Dataset/measurement constraints (generalizability, missingness, transcription)
\paragraph{Dataset- and measurement-specific constraints.}
Our empirical demonstrations are grounded in a single intensive longitudinal cohort (PTSD-STOP) with daily self-reports and diary entries, which enables the methodological analyses we pursue but also constrains external validity.
First, the cohort’s demographics and recruitment context (a clinically monitored trauma-exposed population) may not represent other clinical groups or the general population, and prior work has cautioned that clinical NLP systems can inherit and amplify such representational skews if generalization targets and subgroup performance are not made explicit \citep{bear-dont-walk-etal-2022-ethics-clinical-nlp,obermeyer-etal-2019-racial-bias}.
Second, intensive longitudinal data are rarely missing-at-random: adherence varies over time and across individuals, and missingness mechanisms can bias both modeling and evaluation if not carefully characterized \citep{shiffman-etal-2008-ema,stone-etal-2023-ema-pressing-issues,heron-etal-2017-ema-youth}.
Third, we rely on automatic speech recognition to obtain transcripts; transcription errors and diarization artifacts can introduce additional noise that may differentially affect within-person versus between-person signals, and should be treated as part of the measurement process rather than ignored \citep{shiffman-etal-2008-ema,stone-etal-2023-ema-pressing-issues}.

\section*{Ethical Considerations}

% Ethics Para 1: Human subjects / privacy / governance (HIPAA, consent, provenance documentation)
\paragraph{Sensitive human-subject data and governance.}
This work analyzes longitudinal materials linked to mental health symptom reports, which are inherently sensitive and may contain personally identifying information (even after transcription).
Accordingly, analyses should be conducted under appropriate human-subject oversight and data-use agreements consistent with established principles for research with human participants \citep{belmont-report}.
Because de-identification is not a guarantee of anonymity, especially for rich narrative data, we follow the conservative stance that access controls and minimization of shared artifacts (e.g., avoiding release of raw text/video) are often necessary for participant protection \citep{hhs-hipaa-deid-2025,ayers-etal-2018-dont-quote-me}.
We also view dataset provenance and documentation as part of ethical reporting in NLP: clearly stating who produced the data (sources), under what conditions, and what the intended generalization target is helps prevent misuse and misinterpretation of results \citep{bender-friedman-2018-data,gebru-etal-2018-datasheets}.

% Ethics Para 2: Harms of mental health prediction (stigma, misinterpretation, surveillance; guidance from MH-NLP literature)
\paragraph{Risks of downstream use in mental health contexts.}
Models that predict mental health–related outcomes can be misapplied in ways that harm individuals: erroneous inferences may contribute to stigma, inappropriate monitoring, or decisions made without clinical context.
Prior work in digital mental health NLP emphasizes that prediction should not be conflated with diagnosis, that construct validity is difficult to establish from language traces alone, and that failure modes are often context-dependent \citep{chancellor-dechoudhury-2020-npj-review,ernala-etal-2019-methodological-gaps}.
More broadly, the field has called for explicit ethics disclosures and clearer articulation of intended use, beneficiaries, and plausible harms in mental health prediction research \citep{ajmani-etal-2023-ethics-disclosures,chancellor-etal-2019-taxonomy}.
Consistent with this, our framing and experiments are aimed at improving \emph{methodology} (what we can claim from evaluation), not at promoting clinical deployment.

% Ethics Para 3: Re-identification, quoting, and data leakage
\paragraph{Re-identification and leakage considerations.}
Longitudinal diaries are especially vulnerable to re-identification because they accumulate unique life details over time.
Even when direct identifiers are removed, verbatim excerpts can enable reverse identification, and model artifacts can inadvertently memorize or surface sensitive details \citep{ayers-etal-2018-dont-quote-me,chancellor-etal-2019-taxonomy}.
For this reason, releasing raw text, audio, or video should be treated as a high-risk action; safer alternatives include releasing code, evaluation splits, and aggregate statistics, or providing controlled access mechanisms where appropriate \citep{hhs-hipaa-deid-2025,gebru-etal-2018-datasheets}.

% Ethics Para 4: Fairness, subgroup validity, and clinical safety
\paragraph{Fairness, subgroup validity, and clinical safety.}
Clinical NLP systems can encode uneven performance across demographic or clinical subgroups, potentially exacerbating existing inequities if used in practice \citep{bear-dont-walk-etal-2022-ethics-clinical-nlp,obermeyer-etal-2019-racial-bias}.
A practical implication for longitudinal NLP is that “ecologically valid” splits should be complemented with subgroup-aware reporting whenever sample sizes permit, and that claims should be scoped to the population represented by the data \citep{mitchell-etal-2019-model-cards,bender-friedman-2018-data}.
Finally, mental health applications require particular caution: as emphasized by the CLPsych community, ethical deployment demands careful consideration of consent, stakeholder involvement, and clinical safety boundaries beyond standard ML evaluation \citep{orr-etal-2022-ethical-role}.

\paragraph{Sensitive human-subject data and governance.}
This work analyzes longitudinal materials linked to mental health symptom reports, which are inherently sensitive and may contain personally identifying information (even after transcription).
Accordingly, analyses should be conducted under appropriate human-subject oversight and data-use agreements consistent with established principles for research with human participants \citep{belmont-report}.
The study protocol was reviewed and approved by the relevant Institutional Review Board (IRB), and all researchers followed institutional Human Subjects Research guidelines for secure data handling (e.g., restricted access, secured storage/compute, and minimization of data movement and shared artifacts).
Because de-identification is not a guarantee of anonymity, especially for rich narrative data, we follow the conservative stance that access controls and minimization of shared artifacts (e.g., avoiding release of raw text/video) are often necessary for participant protection \citep{hhs-hipaa-deid-2025,ayers-etal-2018-dont-quote-me}.
% We also view dataset provenance and documentation as part of ethical reporting in NLP: clearly stating who produced the data (sources), under what conditions, and what the intended generalization target is helps prevent misuse and misinterpretation of results \citep{bender-friedman-2018-data,gebru-etal-2018-datasheets}.

\paragraph{Use of AI assistance.}
We used AI assistants to support drafting (e.g., paraphrasing for clarity) and coding/formatting during manuscript preparation.
All generated text and code were reviewed and verified by the authors, and all co-authors participated in editing and auditing to ensure accuracy, appropriate attribution, and consistency with the study protocol and results.

% \section*{Acknowledgments}

% Bibliography entries for the entire Anthology, followed by custom entries
%\bibliography{anthology,custom}
% Custom bibliography entries only
\bibliography{custom, anthology}

\appendix

\section{PTSD-STOP Daily Diary Study}
\label{appsec:ptsd_stop_data}

\paragraph{Open Ended Questionnaire}
Each day, participants completed a brief video diary consisting of 13 open-ended prompts about their day. The camera began recording automatically and the recording uploaded automatically upon submission. Participants were instructed to answer each prompt out loud while facing the camera, to expand on their responses (rather than giving short answers), and to avoid reading the questions aloud. The submit button appeared after 3 minutes, with up to 10 minutes available per prompt. The 13 prompts were:

\begin{enumerate}
\item Tell me about the best part of your day.
\item Tell me about the worst part of your day.
\item Describe when you felt most sad today.
\item Describe when you felt most scared or nervous today.
\item Describe when you felt most annoyed today.
\item Describe when you felt most happy today.
\item How did you feel physically today? Did you have any pain, discomfort, or other physical symptoms? Please elaborate.
\item How did you get along with others today? Please elaborate.
\item Did you have any unwanted, disturbing memories of a past stressful experience? Tell me about this.
\item Today, did you avoid anything because it would have made you uncomfortable? Tell me more about what you did and why.
\item Did you feel on guard today? What made you feel this way?
\item Describe anything that cheered you up today. How did it go?
\item What happened today that you can feel thankful for? Tell me about this.
\end{enumerate}

\paragraph{Rating Scales}
In addition to the open-ended diary, participants completed daily self-report rating scales capturing (i) post-stressor/PTSD-related experiences and overall stress that day, and (ii) exposure to common daily stressors. Rating scales were administered in two parts.

\textbf{Part 1 (Symptom and stress ratings).}
Participants rated the following items everyday, keeping their most stressful event in mind. 
Response options were:
\emph{Not at all, A little bit, Moderately, Quite a bit, Extremely}, and \emph{Skip}.
\begin{enumerate}
\item Today, I had repeated, disturbing, and unwanted memories of the stressful experience.
\item Today, I felt very upset because something reminded me of the stressful experience.
\item Today, I avoided memories, thoughts, or feelings related to the stressful experience.
\item Today, I avoided external reminders of the stressful experience (for example, people, places, conversations, activities, objects, or situations).
\item Today, I felt distant or cut off from other people.
\item Today, I had strong negative feelings such as fear, horror, anger, guilt, or shame.
\item Today, I felt jumpy or easily startled.
\item Today, I was ``superalert'' or watchful or on guard.
\item Overall, how stressed did you feel today?
\end{enumerate}

\textbf{Part 2 (Daily stressor checklist).}
Participants then indicated which of the following troublesome or stressful events occurred everyday (check all that apply):
\begin{enumerate}
\item Had tension or argument with spouse, partner or close family
\item Had tension or argument with others (e.g., co-worker, friend, etc)
\item A lot of demands at home
\item A lot of demands at job
\item A lot of demands made by family
\item Caring for a sick family member
\item Problems with transportation
\item Financial or money problem(s)
\item Health-related event(s)
\item Other troublesome things happened to me
\item No troublesome or stressful things happened to me today
\end{enumerate}

\section{Data Pre-processing}

\subsection{Longitudinal Evaluation}
\label{appsec:preprocess_long_eval}

The violations of independence assumption arise whenever many documents are produced or labeled by a limited set of humans, and they are amplified in longitudinal data where documents are also ordered in time.
Such dependencies are more often the rule than the exception. 
Even when an NLP task may appear on its surface to be ``about the text itself'' (e.g., classification, translation, summarization), supervision originates from a finite set of human sources: annotators create labels, and translators/summarizers create references.
In practice, the ``human set'' can be surprisingly small relative to the number of documents, creating a deeply shared variance structure across instances.

For example, crowdsourced datasets often show extreme annotator imbalances: in the NLU datasets analyzed by \citet{geva-etal-2019-modeling}, MNLI contains 402k examples from 380 annotators and OpenBookQA contains 5k examples from 84 annotators, with the most prolific OpenBookQA annotator contributing 24\% of all examples.
Likewise, widely used summarization corpora contain reference summaries written within a limited set of institutions: Newsroom provides 1.3M article -- summary pairs written by authors and editors across 38 news publications \citep{grusky-etal-2018-newsroom}.
% , and XSum contains 220k BBC articles whose single-sentence summaries are professionally written, typically by the article author \citep{narayan-etal-2018-dont}.
% At the other extreme, many ``user-level'' datasets include tens of thousands of people, but are commonly used in time-aggregated ways~\citep{schwartz-etal-2013-plosone}.
%This mixture of many examples but few human sources, or many sources but little modeled temporal structure, makes it easy for i.i.d.\ document-level evaluation to inadvertently measure source-specific regularities rather than the intended generalization target.

\paragraph{Demonstration of ecological fallacy in traditional evaluation.}
Analyses for Table~\ref{tab:ecovalid} and Figure~\ref{fig:metrics_fig} use the dataset constructed as follows.
To mirror common settings where many documents come from a limited number of human sources \citep{geva-etal-2019-modeling}, we form a controlled subset of 20 individuals from PTSD-STOP.

We first filtered to individuals with non-trivial outcome variation, requiring at least $0.01$ standard deviation in PCL over the full 90-day period, as well as within the first 60 days and last 30 days.
We then sorted individuals by their mean PCL (averaged over days), binned them into 10 strata, and sampled two individuals uniformly at random from each stratum to obtain a balanced cohort of 20 people.

We split this cohort into train/test in three ways:
(a) a document-level random split (70/30), which is ecologically implausible for person-indexed data;
(b) a cross-sectional split (70/30) by person, which holds out individuals; and
(c) a prospective split (70/30) by time, using a temporal cutoff at day $\tau=63$ (70\% of 90 days).
To match train/test sizes across these three settings, we randomly masked a small number of person-day instances.

\paragraph{Between- and within-person metrics isolate what drives performance.}
Analyses for Figure~\ref{fig:metrics_table} use the same 20-person cohort as above, but evaluate one fixed model across multiple test regimes.
We set the prospective temporal cutoff to a 67/33 split (i.e., $\tau \approx 60$) to define the training window, and we additionally construct a cross-sectional \& prospective test set that requires generalization over both people and time.

Concretely, we train a single model on the training region (yellow cells in the split illustration) and evaluate that same model under three regimes: cross-sectional (held-out people), prospective (held-out future days), and cross-sectional \& prospective (held-out people at held-out future days).
This design ensures differences in performance reflect the evaluation regime rather than training different models.

\subsection{Longitudinal Modeling}
\label{appsec:preprocess_long_model}

For all results in \S\ref{sec:modeling}, we use the full analytic sample of 238 participants (Table~\ref{tab:dataset_stats}).
For cross-sectional evaluation, participants are stratified into train/test (80/20) based on their mean PCL over the study period.
For prospective evaluation, we use a fixed temporal cutoff at day $\tau=60$ for training, with later days reserved for testing.

\section{Model Training}
\label{appsec:model_training}

\subsection{Longitudinal Evaluation}
\label{appsec:training_long_eval}

\paragraph{Typical model.}
Our \emph{typical} model is fine-tuning of the task-specific ridge regression layer of a RoBERTa-large encoder~\citep{liu-etal-2019-roberta}. 
We use this term because fine-tuned encoder models remain a strong and widely used baseline for psychological and behavioral prediction from language~\citep{kjell2022natural}, and recent evidence suggests that instruction-tuned LLMs do not consistently outperform fine-tuning smaller encoder models for behavioral prediction and psychological measurement~\citep{singh-etal-2025-systematic, v-ganesan-etal-2023-systematic, choi-etal-2023-llms}.

In practice, we train ridge regression models on transcript representations extracted from the second-to-last layer of RoBERTa-large using DLATK~\citep{schwartz-etal-2017-dlatk}.
Language dimensions were reduced using Principal Component Analysis whenever a hidden dimension size of $le$ 1024 was used.
The ridge penalty (L2 regularization strength) is selected from $\{10^i : i \in [-2,5]\}$ based on performance on a development set.
To avoid leakage, the development set is split from the training data using the same evaluation regime as the corresponding test set (cross-sectional, prospective, or cross-sectional \& prospective).

\subsection{Longitudinal Modeling}
\label{appsec:training_long_model}

\paragraph{Autoregressive ridge models.}
Autoregressive ridge models take as input a length-$h$ history of daily language representations and predict the target outcome (nowcasting or one-day-ahead forecasting, depending on the experiment).
These models are parameterized by history length $h$ and per-day representation size $d$, yielding an input dimension of $h \cdot d$.
Training and hyperparameter selection follow the same procedure as above (ridge regression with the L2 penalty chosen on a regime-matched development split).

\paragraph{Bag-of-embeddings models.} Bag-of-embeddings (BoE) models aggregate a length-$h$ history by averaging the $h$ daily representations into a single vector, which is then used for ridge regression.
As a result, BoE is parameterized by history length $h$ and per-day representation size $d$, but its number of learned parameters scales only with $d$ (since the history is pooled before prediction).
We train BoE models with the same ridge procedure and select the L2 penalty on a regime-matched development split.

\paragraph{Transformers.}
\textbf{Model architecture.}
To test whether modeling \emph{temporal interactions} improves prediction beyond pooled summaries (BoE) and linear dynamics (AR), we trained a minimal autoregressive transformer over the history sequence.
We project each per-day representation from $d{=}128$ down to $d'{=}32$ using a learned linear map, and feed the resulting sequence into a transformer encoder with 1 layer and 1 attention head (hidden size $32$).

For comparability with AR and BoE, we made two design choices:
(1) \emph{No positional embeddings.} We omit positional embeddings to reduce explicit order information, isolating gains attributable to interaction modeling rather than architectural encodings of dynamics; prior work also suggests transformers can recover positional information even without explicit embeddings~\citep{haviv-etal-2022-transformer, Kazemnejad-etal-2023-impact}.
(2) \emph{Fixed history masking.} For a given history length $h$, we apply a lower-triangular attention mask and additionally prevent attention to tokens more than $h$ days in the past by setting the corresponding attention logits to $-\infty$ prior to softmax.
This ensures each transformer configuration is trained and evaluated under the same effective history constraint as the corresponding AR/BoE setting.

\textbf{Hyperparameter optimization.}
We tuned the projection size $d'$, L2 weight decay, attention dropout, output-layer dropout, and learning rate on a regime-matched development split.
We searched projection sizes among powers of two below 128 and selected $d'{=}32$.
The selected hyperparameters were: weight decay $=1$, attention dropout $=0.3$, output dropout $=0.1$, and learning rate $=10^{-3}$ with AdamW.

\textbf{Additional checks.}
Although evaluating full-capacity transformers (e.g., larger depth/width, unmasked attention, or explicit sinusoidal/rotary positional embeddings~\citep{vaswani-etal-attention-2017, su2024roformer}) is beyond our scope, we verified that adding positional embeddings did not yield consistent improvements over our minimal configuration, and that increasing the maximum training history produced the same qualitative patterns reported in Figure~\ref{fig:mae_vs_history_ARBoETrns}.

\section{Metric Functions}
\label{appsec:metrics_fns}
We optimize models using the standard (flattened) mean squared error, leaving alternative objectives and metric formulations for future work.
For evaluation, we report between- and within-person variants of Mean Absolute Error (MAE), Symmetric Mean Absolute Percentage Error (SMAPE), and Pearson correlation ($r$).
SMAPE and $r$ are bounded and scale-invariant, ranging in $[0,2]$ and $[-1,1]$, respectively.

\smallskip
\noindent\textbf{SMAPE.}
For predictions $\hat{y}$ and targets $y$, we compute
\begin{equation}
\mathrm{SMAPE}(y,\hat{y}) \;=\; \frac{2}{N}\sum_{n=1}^{N}
\frac{\left| \hat{y}_n - y_n \right|}
{\left|y_n\right| + \left|\hat{y}_n\right| + \epsilon},
\end{equation}
where $N$ is the number of evaluated instances and $\epsilon$ is a small constant to avoid division by zero.
Since the outcomes range between [1, 5], $\epsilon$ was set to 0.

\smallskip
\noindent\textbf{Pearson correlation.}
We compute Pearson correlation between $y$ and $\hat{y}$ as
\begin{equation}
r(y,\hat{y}) \;=\;
\frac{\sum_{n=1}^{N}\left(y_n-\bar{y}\right)\left(\hat{y}_n-\bar{\hat{y}}\right)}
{\sqrt{\sum_{n=1}^{N}\left(y_n-\bar{y}\right)^2}\;
\sqrt{\sum_{n=1}^{N}\left(\hat{y}_n-\bar{\hat{y}}\right)^2}},
\end{equation}
where $\bar{y}=\frac{1}{N}\sum_{n=1}^{N}y_n$ and $\bar{\hat{y}}=\frac{1}{N}\sum_{n=1}^{N}\hat{y}_n$.

\begin{figure}
\centering
\includegraphics[width=\linewidth]{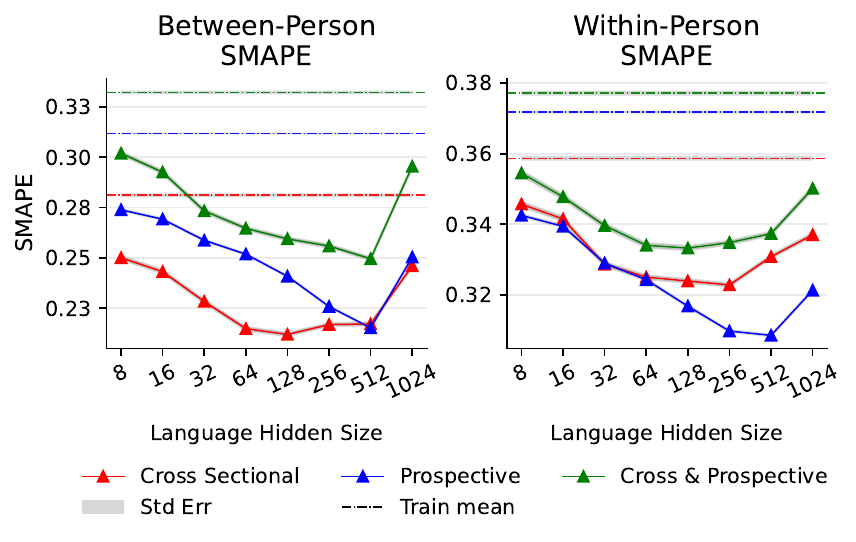}
\caption{\textbf{Between- \textit{(top)} and Within-Person \textit{(bottom)} SMAPE as a function of hidden dimension size.} Forecasting performance follows a U-shaped trend as a function of hidden dimension size of language across all three evaluation sets.  
While a typical model requires only 64 dimensions of language for best performance on Cross-sectional and Cross-sectional \& Prospective test sets, it requires 512 dimensions in Prospective evaluation set.  
Based on the best performance achieved in different settings, generalization to Cross-sectional \& Prospective is the hardest, followed by Cross-sectional set and finally prospective set. 
Generalization to unseen people is harder than unseen time and within-person changes is harder than between-person differences.  
}
\label{fig:smape_vs_hdnsize}
\end{figure}

\begin{figure*}
\centering
\includegraphics[width=\linewidth]{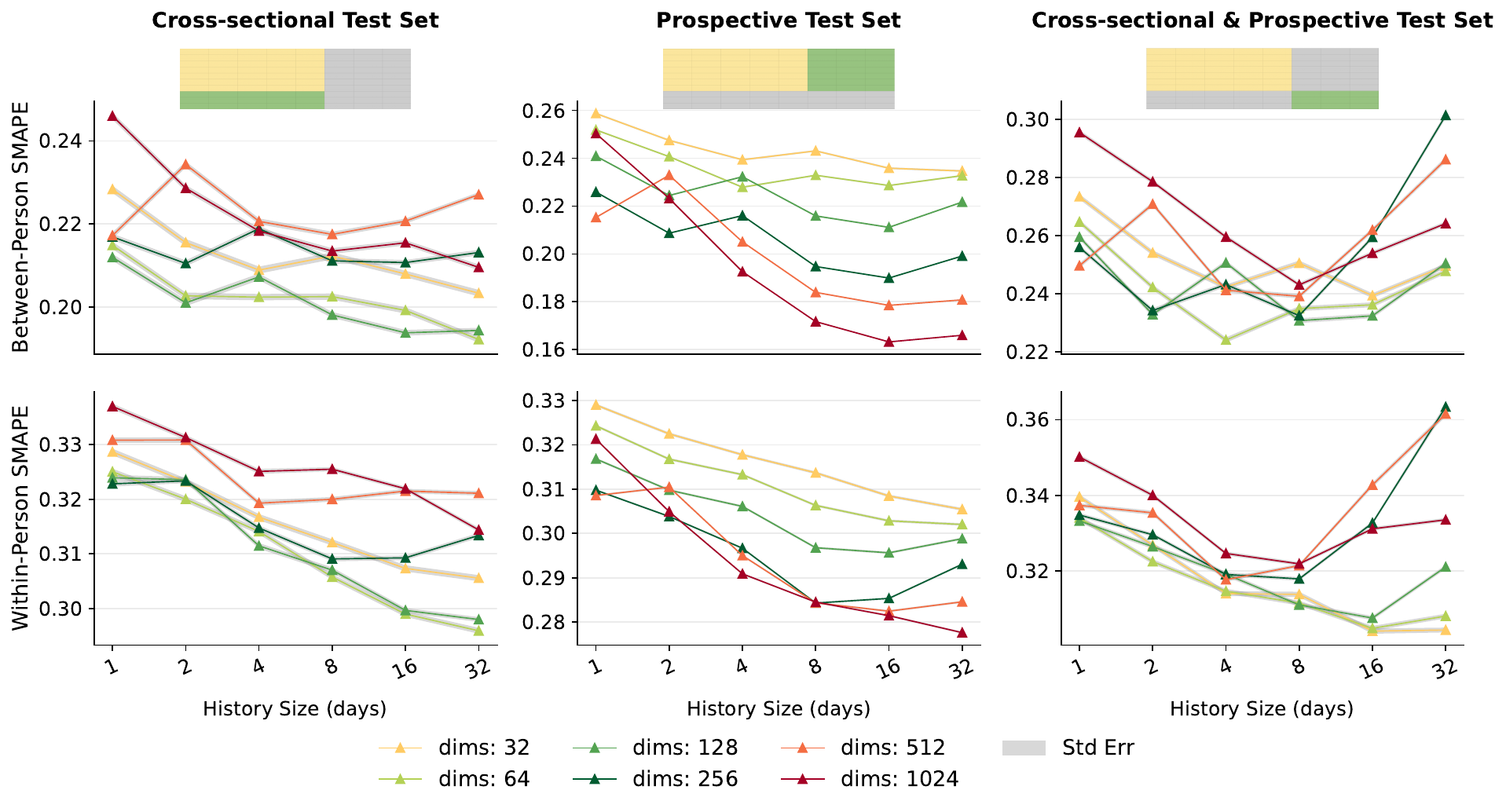}
\caption{\textbf{Between- and Within-Person SMAPE as a function of history length for Auto Regressive Model.}
Predictive Performance improves with modeling the temporal dynamics of linguistic behavior. 
For Prospective test set, between- and within-person performance improves with longer temporal context and higher dimensional sizes (512 and 1024).  
For Cross-sectional and Cross-sectional \& Prospective evaluation sets, performance improves with temporal context at lower dimensions (hidden size=64).
}
\label{fig:smape_vs_history}
\end{figure*}

\begin{figure*}
\centering
\begin{subfigure}[T]{\textwidth}
    \caption{\textbf{Between- \textit{(top)} and Within-Person \textit{(bottom)} MAE as a function of history length for Bag of Embeddings Model.}}
    \label{fig:mae_vs_history_boe}
    \includegraphics[width=\linewidth]{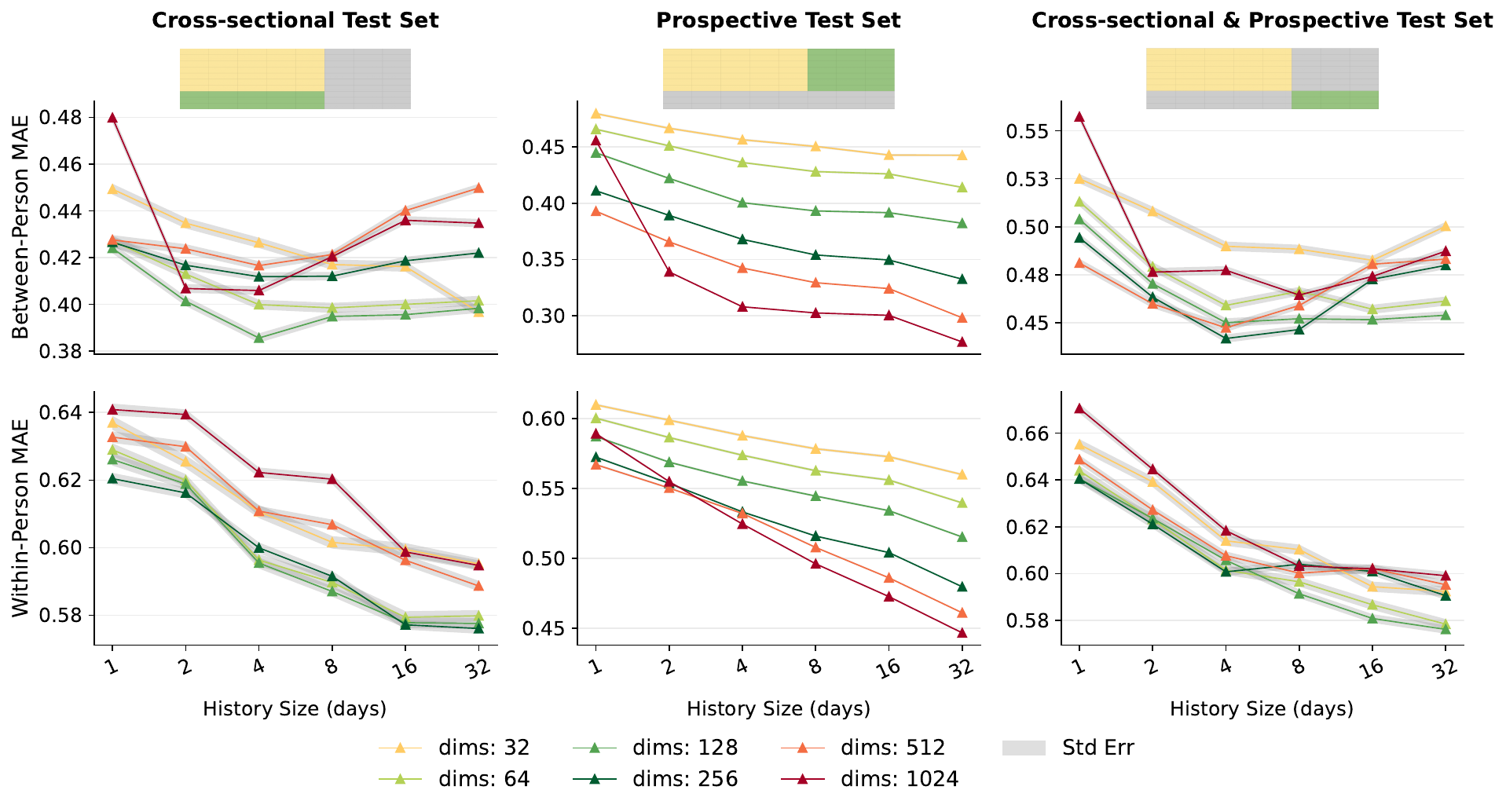}
\end{subfigure}
\hfill
\begin{subfigure}[T]{\textwidth}
    \caption{\textbf{Between- \textit{(top)} and Within-Person \textit{(bottom)} SMAPE as a function of history length for Bag of Embeddings Model.}}
    \label{fig:smape_vs_history_boe}
    \includegraphics[width=\linewidth]{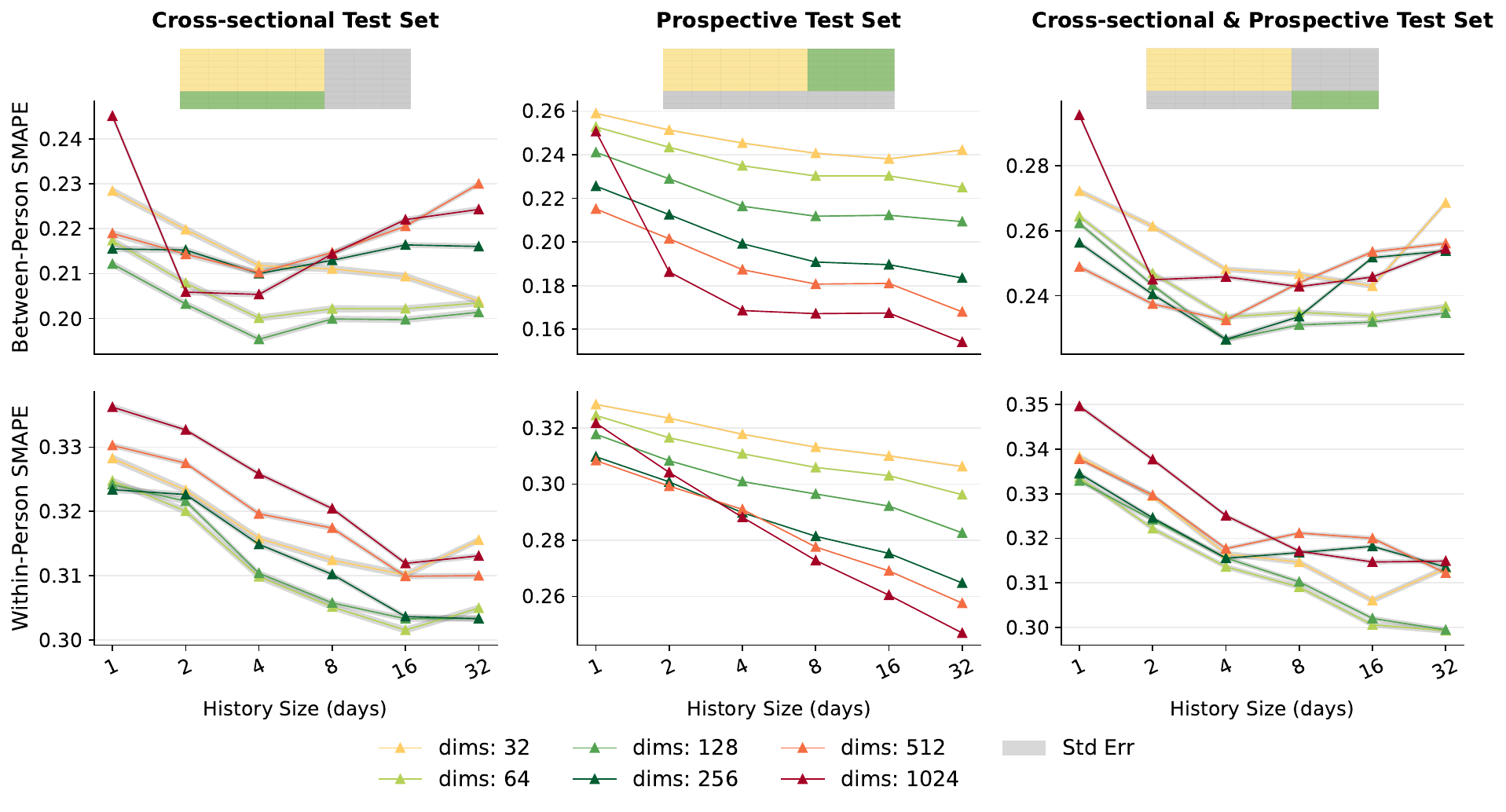}
\end{subfigure}
\caption{\textbf{Between- and Within-Person metrics as a function of history length for Bag of Embeddings Model.}
Predictive Performance improves with modeling the temporal context of linguistic behavior. 
For Prospective test set, between- and within-person performance improves with longer temporal context and higher dimensional sizes (512 and 1024).  
For Cross-sectional and Cross-sectional \& Prospective evaluation sets, performance improves with temporal context at lower dimensions (hidden size=64-128).}
\end{figure*}

\begin{figure*}
\caption{\textbf{Comparison of Auto Regressive, Bag of Embeddings and Transformer model using Between- \textit{(top)} and Within-Person \textit{(bottom)} SMAPE as a function of history length.}}

\includegraphics[width=\linewidth]{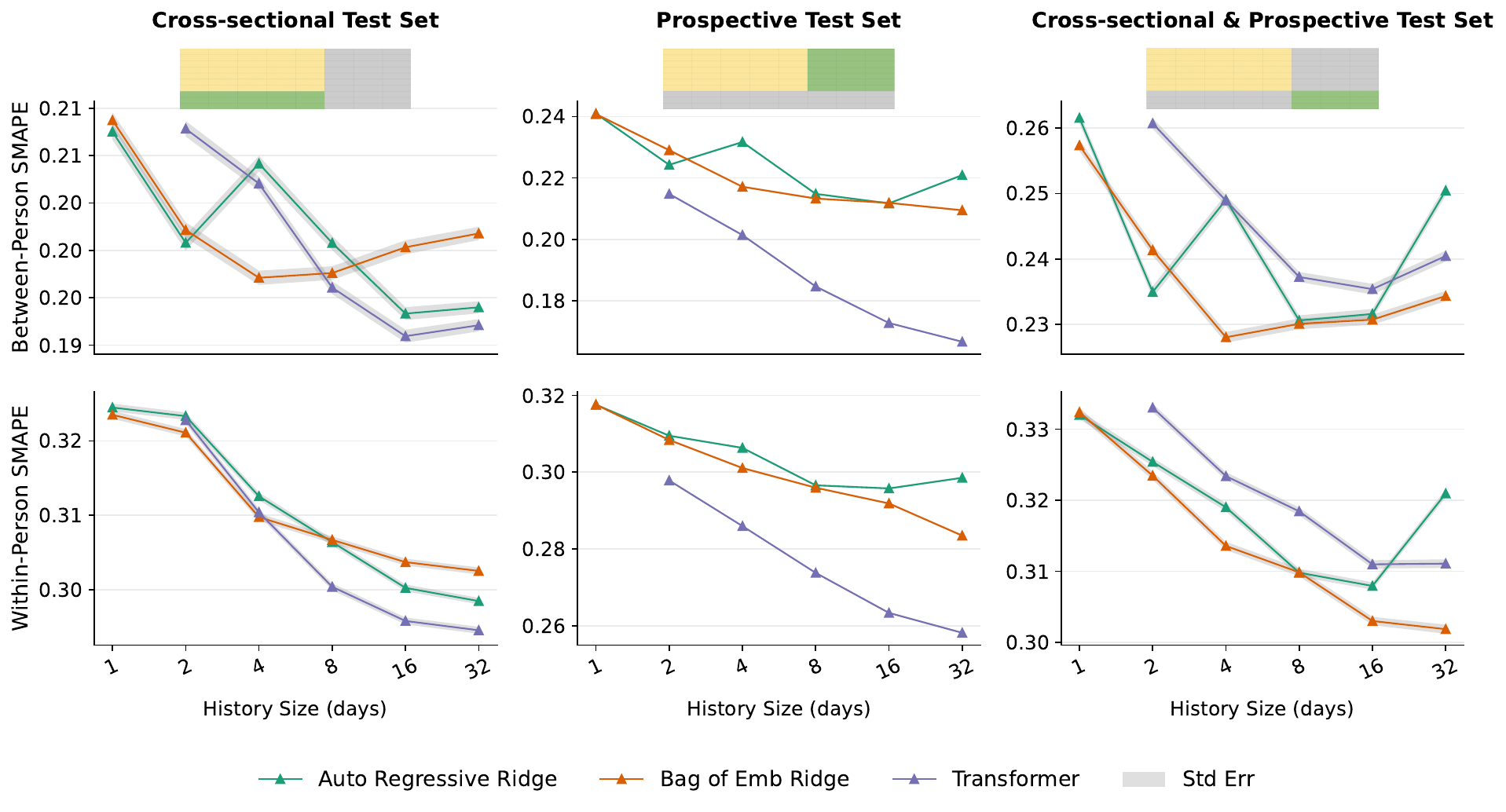}
\caption{\textbf{AR vs.\ BoE vs.\ Transformer across history length (128 dims per day).}
Between-person (top) and within-person (bottom) SMAPE as a function of history $h$.
Modeling temporal interactions (Transformer) offers marginal improvements over modeling dynamics in cross-sectional splits, but yields the largest gains for prospective generalization. It performs worst for cross-sectional \& prospective.}
\label{fig:smape_vs_history_ARBoETrns}
\end{figure*}

\end{document}